\documentclass{article}
\usepackage{spconf,amsmath,graphicx}
\usepackage{lgrind}
\usepackage{graphicx,amsmath,bm,amsfonts,enumitem,amssymb,subfigure,url,listings}
\usepackage{multirow}
\usepackage{booktabs}
\usepackage{tikz}
\usepackage{breqn}
\usetikzlibrary{shapes.geometric, shapes.misc, arrows, arrows.meta, positioning}
\graphicspath{figures/}

\def\L{{\cal L}}

\title{Neural Network Pruning Through Constrained Reinforcement Learning}

\name{Shehryar Malik, Muhammad Umair Haider, Omer Iqbal, Murtaza Taj}
\address{ LUMS School of Science and Engineering}
\begin{document}
%
\maketitle
\begin{abstract}
 Network pruning reduces the size of neural networks by removing (pruning) neurons such that the performance drop is minimal. Traditional pruning approaches focus on designing metrics to quantify the usefulness of a neuron which is often quite tedious and sub-optimal. More recent approaches have instead focused on training auxiliary networks to automatically learn how useful each neuron is however, they often do not take computational limitations into account. In this work, we propose a general methodology for pruning neural networks. Our proposed methodology can prune neural networks to respect pre-defined computational budgets on arbitrary, possibly non-differentiable, functions. Furthermore, we only assume the ability to be able to evaluate these functions for different inputs, and hence they do not need to be fully specified beforehand. We achieve this by proposing a novel pruning strategy via constrained reinforcement learning algorithms. We prove the effectiveness of our approach via comparison with state-of-the-art methods on standard image classification datasets. Specifically, we reduce $83-92.90\%$ of total parameters on various variants of VGG while achieving comparable or better performance than that of original networks. We also achieved $75.09\%$ reduction in parameters on ResNet18 without incurring any loss in accuracy. 
 
\end{abstract}

\section{Introduction}
\label{sec:intro}
Deep neural networks typically have large memory and compute requirements, making it difficult to deploy them on small devices such as mobiles and tablets~\cite{KimICLR2016}. These cumbersome networks are pruned by removing their redundant weights, layers, filters and blocks~\cite{Tan2019EfficientNetRM, he2018amc, lemaire2019structured, YanMIPR2020, davis2020state, UmairICIP2021}. Neural network pruning strategies can be grouped into three categories namely i) Offline pruning, ii) Online pruning and iii) Pruning via Reinforcement Learning.

\textbf{Offline Pruning} requires multiple iterations of training, pruning, and fine-tuning. magnitude-based pruning~\cite{Han2015nips} which works on `magnitude equals salience' principle. \cite{lecun1990obd, LebedevCVPR2016} uses second derivative of weights to rank connections, More recent approaches suggest look-ahead pruning strategies~\cite{lee2021lookahead, YanMIPR2020} where magnitude of previous connected layers is also considered. Layers are pruned independently to speed up the pruning process in ~\cite{dong2017layerobd}. Instead of reducing the original, deep network (called the ``teacher'') by pruning connections, knowledge distillation~\cite{hinton2015distilling} trains another, smaller network (called the ``student'') to mimic the output of the teacher network. However, it requires designing an ad hoc student network which is a tedious task.\\

\textbf{Online Pruning}: A more recent class of techniques poses the problem of pruning as a learning problem by introducing a mask vector that acts as a gate or an indicator function to turn on/off a particular component (connection, filter, layer, block)~\cite{li2017filter, HuangECCV2018, ZhanhongPCNNDAC2020, UmairICIP2021}. The mask vector can be treated as a trainable parameter and is directly optimized through gradient descent. The mask can also be obtained via an auxiliary neural network~\cite{lin2017runtime,he2018amc}. Pruning via Surrogate Lagrangian Relaxation (P-SLR)~\cite{GurevinArxiv2020} utilizing surrogate gradient algorithm for Lagrangian Relaxation~\cite{ZhaoSLR1997}. ~\cite{lemaire2019structured, Enderich_2021_WACV, Tan2019EfficientNetRM,Enderich_2021_WACV} uses Budget Aware Pruning hat tries to reduce the size of a network to respect a specific budget on computational and space complexitie these budgets introduce an arbitrary function that is non-differentiable. However, one shortcoming of current budget-aware techniques is that they require the function on which the budget is being imposed to be either differentiable or fully specified~\cite{lemaire2019structured}. This, however, is not always possible when we, for example, want to impose budgets on metrics such as inference time. The main question, then, is the following: can we prune neural networks to respect budgets on arbitrary, possibly non-differentiable, functions? One way to solve this problem is to leverage some recent techniques in reinforcement learning (RL)~\cite{sutton1998rl,reinforcementlearning}.

\textbf{Pruning via Reinforcement Learning}: In reinforcement learning~\cite{reinforcementlearning} an agent interacts with the environment around them by taking different actions. Each action results in the agent receiving a reward depending upon how good or bad the action is. If the agent is trained to predict sparsity as action, then the accuracy of the obtained model can be used as a reward, resulting in RL based neural network pruning~\cite{he2018amc}. Reinforcement learning (RL)~\cite{sutton1998rl,reinforcementlearning} is known to optimize arbitrary non-differentiable functions while respecting budget on computational and space complexities. One such example is AutoML for Model Compression (AMC)~\cite{he2018amc} in which an agent is trained to predict the sparsity ratio for each layer. Accuracy is returned as a reward to encourage the agent to build smaller, faster, and more accurate networks. Similarly, Conditional Automated Channel Pruning (CACP)~\cite{CACP2021IEEESPL} is another RL-based method that simultaneously produces compressed models under different compression rates through a single layer-by-layer channel pruning process. While these methods allow to include both sparsity and accuracy, they lack fine-grained control over these constraints. 

\textbf{Our main contributions}: In this work, we formulate the problem of pruning in the constrained reinforcement learning framework (CRL). We propose a general methodology for pruning neural networks, which can prune neural networks to respect specified budgets on arbitrary, possibly non-differentiable, functions. Furthermore, we only assume the ability to be able to evaluate these functions for different inputs, and hence they do not need to be fully specified beforehand. Our proposed CRL framework outperform state-of-the-art methods in compressing popular models such as VGG~\cite{simonyan2015vgg} and ResNets~\cite{he2015resnet} on benchmark datasets.

%


\section{Pruning via Constrained Reinforcement Learning}
\label{sec:method}


Constrained reinforcement learning~\cite{altman-constrainedMDP} is an extension of the reinforcement learning problem in which agents, in addition to the reward, also receive a cost. The agent's goal is to maximize its cumulative reward subject to its cumulative cost being less than some pre-defined threshold. One interesting property here is that neither the reward nor the cost function needs to be differentiable or fully specified. The agent simply needs to be fed a scalar reward and a scalar cost value each time it performs an action.

Let $\boldmath\theta$ denote the parameters of a neural network. Each element of $\boldmath\theta$ represents the weight of a single connection. Removing a connection is thus equivalent to multiplying its weight by $0$. Let $\mathcal{L}$ denote the loss function of the neural network. Pruning, in its most general form, tries to find a mask $\boldmath{M} \in \{0,1\}^{\vert\theta\vert}$ that solves the following optimization:
\begin{equation}
\begin{split}
&\underset{\boldmath\theta,\boldmath{M}\in\{0,1\}^{\vert\theta\vert}}{\text{minimize}}\; \mathcal{L}(\boldmath\theta\odot \boldmath{M})\\
&\text{subject to } f(\boldmath{M}) \leq \alpha,
\end{split}
\end{equation}
where $\odot$ is the element-wise Hadamard product, $f$ be some arbitrary, possibly non-differentiable function, and $\alpha$ is a known constant. Here $f$ represents the computational and space complexity of the network.  For example, if we want to compress the network by at least $50\%$ (in terms of the space it occupies), we can let $f$ be the $\ell_1$-norm and $\alpha$ to be equal to $0.5\vert\boldmath\theta\vert$. Similarly, if we wanted to optimize for speed, we could set $f$ to be the number of flops the network consumes (or the time it takes) when it is pruned according to $\boldmath{M}$.  

\begin{figure}[t!]
	\centering
	\includegraphics[width=1\columnwidth]{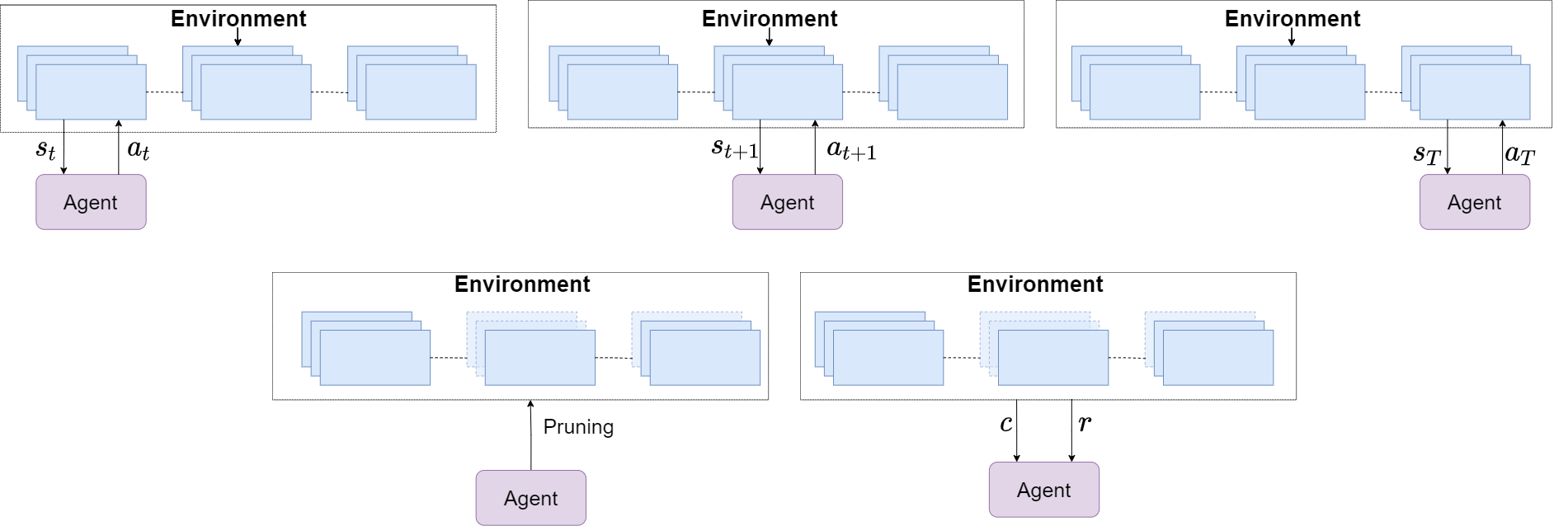}
	\caption{A graphical visualization of the environment consisting of a pre-trained network that we are interested in pruning. At time-step $t$, the policy outputs an action that specifies how the $t^{th}$ convolutional layer should be pruned. The environment executes this action and returns the next state (i.e., layer $(t+1)^{th}$ specifications) and a reward and a cost.}
	\label{fig:env}
\end{figure}

In our proposed approach, the mask $\boldmath{M}$ is modeled via action performed by an agent who is interacting with the environment. In the beginning, the environment loads a pre-trained network (e.g., VGG16~\cite{simonyan2015vgg}). The agent is then fed the layer representations of the first convolutional layer in the network (as defined in the previous section). The agent then specifies an appropriate action. This action is then recorded, and the agent is then fed the next state. Once the agent has finished predicting its actions for all convolutional layers, the network is pruned according to the agent's specified actions and fine-tuned on the training set (this is the same training set that was used to pre-train the network). The accuracy of the fine-tuned network is returned as the agent's reward along with the cost (the reward and cost at all other time-steps are $0$). Figure~\ref{fig:env} shows an illustration of our pruning process.

\subsection{Constrained Markov Decision Process}
\label{ssec:mdp}

\textbf{Markov Decision Processes}
Mathematically, the notion of an environment is captured through a Markov Decision Process (MDP). A finite-horizon MDP $\mathcal{M}$ is a tuple $(\mathcal{S}, \mathcal{A}, p, r, \gamma, T)$, where $\mathcal{S} \in \mathbb{R}^{\vert\mathcal{S}\vert}$ is a set of states, $\mathcal{A}\in \mathbb{R}^{\vert\mathcal{A}\vert}$ is a set of actions, $p:\mathcal{S}\times\mathcal{A}\times\mathcal{S} \mapsto [0,1]$ is the transition probability function (where $p(s'\vert s,a)$ denotes the probability of transitioning to state $s'$ from state $s$ by taking action $a$), $r:\mathcal{S} \times \mathcal{A} \mapsto \mathbb{R}$ is the reward function, $\gamma$ is the discount factor and $T$ is the time-horizon. A trajectory $\tau=\{s_1,a_1,\ldots,s_T,a_T\}$ denotes a sequence of states-action pairs such that $s_{t+1} \sim p(\cdot \vert s_t,a_t)$.

The goal of an agent is to learn a \textit{policy} about which action to take in each state. Formally, a policy $\pi : \mathcal{S} \mapsto \mathcal{P}(\mathcal{A})$ is a map from states to probability distributions over actions, with $\pi(a\vert s)$ denoting the probability of taking action $a$ in state $s$. 

Let $r(\tau) = \sum_{t=1}^T \gamma^t r(s_t,a_t)$ denote the total discounted reward of a trajectory. The problem of RL is to find a policy $\pi^*$ that maximizes the expected total discounted reward, i.e.,
\begin{equation}
\pi^* = \text{arg}\max_{\pi} J(\pi) = \mathbb{E}_{\tau\sim\pi}[r(\tau)].
\label{eq:rl-prob}
\end{equation}
We parameterize $\pi$ as a neural network to deal with large state and action spaces. One common approach is to model $\pi$ as a Gaussian distribution~\cite{NIPS1995_7cce53cf} $\mathcal{N}(\mu_\theta(\boldmath{s}),\boldmath\sigma I)$, where $\mu_\theta :\vert\mathcal{S}\vert\mapsto\vert\mathcal{A}\vert$ is a neural network. $\boldmath\sigma\in\mathcal{R}^{\vert\mathcal{A}\vert}$ is a trainable vector and $I$ is the identity matrix. We will denote all of the parameters of $\pi$ with $\boldmath\theta$.
The policy gradients algorithm~\cite{william1992reinforce} finds $\pi^*$ through stochastic gradient descent. That is, we update $\boldmath\theta$ as
\begin{equation}
\boldmath\theta := \boldmath\theta + \nabla_{\boldmath\theta} J(\pi_{\boldmath\theta}).
\end{equation}
The gradient of loss $J$ with respect to $\boldmath\theta$ can be shown to be
\begin{equation}
\nabla_{\boldmath\theta}J(\boldmath\theta) = \nabla_{\boldmath\theta}\mathbb{E}_{\tau\sim\pi_{\boldmath\theta}}[r(\tau)] = \mathbb{E}_{\tau\sim\pi_{\boldmath\theta}}[\nabla_{\boldmath\theta}\log \pi_\theta(\tau)r(\tau)].
\label{eq:policy-gradient}
\end{equation}
The proximal policy algorithm (PPO)~\cite{schulman2017proximal} makes a first-order approximation to this optimization problem and proposes to update $\pi_{\boldmath\theta}$ for some known constant $\epsilon$ by solving

\begin{dmath}
\pi^* = \text{arg}\max_\pi \mathbb{E}_{\tau\sim\pi_{\bar{\bm\theta}}}(x)
\end{dmath}
\begin{dmath}
x = \left[\min\left(\frac{\pi_{\bm\theta}(\tau)}{\pi_{\bar{\bm\theta}}(\tau)}r(\tau),
\text{clip}\left(\frac{\pi_{\bm\theta}(\tau)}{\pi_{\bar{\bm\theta}}(\tau)},1-\epsilon,1+\epsilon\right)r(\tau)\right)\right]
\end{dmath}
Here $\text{clip}$ bounds its first argument between the other two. The objective function above essentially removes any incentive to move the ratio between the two policies outside of the interval $[1-\epsilon,1+\epsilon]$. 

\textbf{Constrained Markov Decision Processes}: In order to formulate our problem as a constrained RL problem, we need to first define our constrained MDP (CMDP). Let $g = g_F \circ g_T \circ g_{T-1} \circ \ldots g_1$ define a neural network with $T$ convolutional layers, $g_1,\ldots,g_T$ followed by a few fully connected layers collectively denoted as $g_F$. We wish to prune $g$. Furthermore, let $\boldmath\theta_t$ denote the parameters corresponding to layer $t$. We define the key components of the CMDP as follows:
\begin{enumerate}
    \item State, $s$: Each convolutional layer corresponds to a single state. Similar to the scheme in \cite{he2018amc} each of these layers is represented by the following tuple:
    \begin{center}
        (layer ($t$), input channels, number. of filters, kernel size, stride, padding)    
    \end{center}
    where $t$ is the index of that layer and the remaining entries are the attributes of a convolutional layer.
    \item Action, $a$: for the action, representations we end up with a mask vector $M_t$ at each layer $g_t$. The length of this mask vector will be equal to the number of filters for $g_t$. We use $M$ to collectively denote the mask vectors for all layers.
    \item Transition function: The agent always transitions from state $g_t$ to state $g_{t+1}$, and so the transitions are fixed and independent of the agent's actions.
    \item Reward function, $r$: Let $B$ be a batch of input examples (uniformly) sampled from the training dataset. We define the reward as follows:
    \begin{equation}
        r(s_t,a_t) =
        \begin{cases}
            -\mathcal{L}(\theta \odot M) & \text{if } t=T\\
            0 & \text{otherwise}
        \end{cases}
    \end{equation}
    where $\mathcal{L}$ is the loss function of the network evaluated on the batch $B$.
    \item Cost function, $c$: the cost function is defined as
    \begin{equation}
        c(s_t,a_t) =
        \begin{cases}
            f(M) & \text{if } t=T\\
            0 & \text{otherwise}
        \end{cases}
    \end{equation}
where $f$ is our constraint function evaluated on the batch $B$.
    \item Budget, $\alpha$: This the budget on $f$ we wish our pruned network to respect. 
\end{enumerate}
The \textbf{policy} predicts a sparsity ratio for each layer. Filters are then pruned via magnitude-based pruning~\cite{Han2015nips} upto the desired sparsity.

\subsection{Algorithm}
\label{ssec:algo}

Let $d_a$ be the dimension of the action. Also, let $d_s$ denote the dimension of the state vector (recall that each state vector corresponds to a single layer).

We model our policy as a (diagonal) multivariate Gaussian distribution $\mathcal{N}(\mu_{\boldmath{\theta}}, \boldmath{\sigma} \boldmath{I})$. Here $\mu_{\boldmath{\theta}}:\mathbb{R}^{d_s}\mapsto\mathbb{R}^{d_a}$ is a neural network with parameters $\boldmath{\theta}$, $\sigma\in\mathbb{R}^{d_a}$ is a trainable vector and $\boldmath{I}\in\mathbb{R}^{d_a\times d_a}$ is the identity matrix (hence $\boldmath{\sigma} \boldmath{I}$ is the covariance matrix of the distribution). The network $\mu_{\boldmath{\theta}}$ takes in as input a state vector $s$ provided by environment and outputs a vector of dimension $d_a$. We then simply sample an action from $\mathcal{N}(\mu_{\boldmath{\theta}}(\boldmath{s}),\boldmath{\sigma}\boldmath{I})$ and feed it to the environment.
In parallel, we also train two other neural networks, $V_{\boldmath{\phi_r}}^r:\mathbb{R}^{d_s}\mapsto\mathbb{R}$ and $V_{\boldmath{\phi_c}}^c:\mathbb{R}^{d_s}\mapsto\mathbb{R}$ with parameters $\boldmath{\phi_r}$ and $\boldmath{\phi_c}$. The cost value function is also defined similarly for the cost function. Recall that having these networks helps us reduce variance.

We initialize our policy and value networks randomly and collect data $\mathcal{D}$ from the environment. Each data point is essentially a tuple $({s}_t,{a_t},{s}_{t+1},r,c)$ where ${s}_t$ and ${s}_{t+1}$ are the states at time $t$ and $t+1$ respectively, ${a}_t$ is the action taken at time $t$ and $r$ and $c$ are the reward and cost received as a consequence of taking action ${a}_t$. We use this dataset to update our parameters. Furthermore, we initialize our Lagrange multiplier $\lambda$ with a constant value and also update it using this dataset. This entire process is repeated until convergence.

Recall that at each iteration, we are interested in optimizing (here we decompose the expectation over trajectories into expectation over states and actions):
%
\begin{dmath}
    J_{\text{LAG}}^{\text{PPO}}(\lambda,\pi_{\boldmath\theta}) = \sum_{t=1}^T \sum_{{s}_t,{a}_t\sim\mathcal{D}}\left[\frac{\pi_{\boldmath\theta}({s}_t,{a}_t)}{\pi_{\bar{\boldmath\theta}}({s}_t,{a}_t)}J^r({s}_t,{a}_t,{s}_{t+1}) - \lambda (\mathbb{E}_{\tau\sim\pi_{\boldmath\theta}}[J^c({s}_t,{a}_t,{s}_{t+1})] - \alpha)\right],
\end{dmath}
%
where $\alpha$ is the budget and the losses $J^{(.)}$ are: 
%
\begin{equation}
    J^r({s}_t,{a}_t,{s}_{t+1}) = \sum_{t'=t}^T r({s}_t,{a}_t) - V_{\boldmath{\phi_r}}^r({s}_{t+1}),
\end{equation}
\begin{equation}
    J^c({s}_t,{a}_t,{s}_{t+1}) = \sum_{t'=t}^T c({s}_t,{a}_t) - V_{\boldmath{\phi_c}}^c({s}_{t+1}).
\end{equation}
All parameters are updated via to gradient descent using their respective learning rates $\eta_i$. Specifically, the policy network is updated as:
\begin{equation}
    \boldmath{\theta} := \boldmath{\theta} - \eta_1\nabla_{\boldmath\theta}J_{\text{LAG}}^{\text{PPO}}(\lambda,\pi_{\boldmath\theta}),
\end{equation}
%
and the Lagrange multiplier as:
%
\begin{equation}
    \lambda := \lambda - \eta_2\nabla_{\lambda}J_{\text{LAG}}^{\text{PPO}}(\lambda,\pi_{\boldmath\theta}),
\end{equation}
%
Furthermore, we define the loss for the reward value function network as:
\begin{equation}
    \mathcal{L}^r = \sum_{t=1}^T \sum_{{s}_t,{a}_t\sim\mathcal{D}}\vert\vert r(s_t,a_t) + \gamma_r V_{\boldmath{\phi_r}}(s_{t+1}) - V_{{\boldmath\phi_r}}(s_t)\vert\vert_2^2,
\end{equation}
%
and update its parameters as:
%
\begin{equation}
    \boldmath{\phi_r} := \boldmath{\phi_r} - \eta_3 \nabla_{\boldmath{\phi_r}}\mathcal{L}^r(\boldmath{\phi_r}),
\end{equation}
%
Similarly, the loss for the cost value function network is defined as:
%
\begin{equation}
    \mathcal{L}^c = \sum_{t=1}^T \sum_{\boldmath{s}_t,\boldmath{a}_t\sim\mathcal{D}}\vert\vert c(s_t,a_t) + \gamma_c V_{\boldmath{\phi_c}}(s_{t+1}) - V_{{\boldmath\phi_c}}(s_t)\vert\vert_2^2,
\end{equation}
and its parameters are updated as:
\begin{equation}
    \boldmath{\phi_c} := \boldmath{\phi_c} - \eta_4\nabla_{\boldmath{\phi_c}}\mathcal{L}^c(\boldmath{\phi_c}),
\end{equation}
\section{Results}
\label{sec:exp}

\subsection{Experimental Setup}
\label{ssec:expsetup}
\vspace{-0.2cm}
We evaluated our approach using CIFAR-10~\cite{krizhevsky2009learning} dataset on ResNet18~\cite{he2015resnet} and variants of VGG~\cite{simonyan2015vgg} network. The training was performed using Adam optimizer~\cite{kingma2014method} using learning rates $\eta_i=3.0\times 10^{-4}$ and batch size $60$. Experiments were conducted on VGG11, VGG16, VGG19, and ResNet18. For each model, we ran the experiments using different budget values. Initially, the policy network was trained for a fixed number of iterations. The policy network then pruned the original model by predicting the sparsity ratio for every convolutional layer using the PPO-Lagrangian algorithm. The Lagrangian multiplier was initialized with a fixed value as $\lambda=1$, and $\gamma_r$ and $\gamma_c$ are initialized as $0.99$ and $1.00$ respectively. Once the network was pruned, it was fine-tuned for a certain number of iterations. Since fine-tuning is computationally expensive, we adopted a schedule with hyperparameter values of 0, 32, 128, respectively. It means that we fine-tune less in the beginning and more towards the end. In practice, we normalize our rewards and cost values with a running mean and standard deviation which is continually updated as more data is collected. Furthermore, we also normalize the state vector in a similar way to improve stability.


\subsection{Ablative Study}
\label{ssec:ablstudy}
\vspace{-0.2cm}
To demonstrate the efficacy of the proposed constrained RL method, we performed experiments on VGG11, VGG16, and VGG19~\cite{simonyan2015vgg} and compared the results with the magnitude-based pruning method. We train each of these networks on the CIFAR-10 dataset~\cite{krizhevsky2009learning}. Pretrained models of VGG were used to train the policy network for $40$K iterations. During pruning, the models were fine-tuned by $25$K, $35$K, $40$K iterations, respectively, in a fine-tuning schedule. We experimented with budget values $\alpha=10$ and $\alpha=20$. Increasing the $\alpha$ value decreases the sparsity of the pruned network. The ablative study showed that the proposed constrained RL method is significantly more optimal than magnitude-based pruning (see Table~\ref{tab:ablstudy}). Our method achieved higher accuracy than magnitude-based pruning in all experiments. In fact, it even outperformed the unpruned network in 4 out of 6 cases.

\begin{table}[t]
  \centering
  \caption{Comparison of sparsity and accuracy between unpruned network, baseline magnitude-based pruning~\cite{Han2015nips} and proposed constrained reinforcement learning (CRL).}
  \resizebox{0.99\columnwidth}{!}{
    \begin{tabular}{rl|cc|cc}
          &       & \multicolumn{2}{c|}{$\alpha=20$} & \multicolumn{2}{c}{$\alpha=10$} \\
          &       & \multicolumn{1}{l}{Sparsity ($\%$)} & \multicolumn{1}{l|}{Acc. ($\%$)} & \multicolumn{1}{l}{Sparsity ($\%$)} & \multicolumn{1}{l}{Acc. ($\%$)} \\
    \midrule
    \multirow{3}[1]{*}{\rotatebox[origin=c]{90}{VGG11}} & Unpruned & "-"   & \textbf{89.23} & "-"   & \textbf{89.23} \\
          & Magnitude-based Pruning & 80.00 & 85.50 & 90.00 & 83.80 \\
          & Proposed CRL & \textbf{83.48} & 89.11 & \textbf{90.75} & {88.09} \\
    \midrule
    \multirow{3}[1]{*}{\rotatebox[origin=c]{90}{VGG16}} & Unpruned & "-"   & 90.69 & "-"   & \textbf{90.69} \\
          & Magnitude-based Pruning & 80.00 & 88.40 & 90.00 & 87.10 \\
          & Proposed CRL & \textbf{83.81} & \textbf{90.96} & \textbf{92.90} & 89.89 \\
    \midrule
    \multirow{3}[1]{*}{\rotatebox[origin=c]{90}{VGG19}} & Unpruned & "-"   & 90.59 & "-"   & 90.59 \\
          & Magnitude-based Pruning & 80.00 & 88.40 & 90.00 & 86.90 \\
          & Proposed CRL-Coarse & \textbf{83.48} & \textbf{91.06} & \textbf{92.31} & \textbf{91.31} \\
    \end{tabular}
   }%
  \label{tab:ablstudy}%
\end{table}%

\subsection{Comparison with State-of-the-art}
\label{ssec:soacomp}
\vspace{-0.2cm}
To prove the effectiveness of the proposed constrained RL approach, we also conducted experiments with ResNet18, VGG16 and Resnet50 to compared our results with state-of-the-art methods on the CIFAR-10 dataset.

In the case of VGG16, we compared our method with two state-of-the-art methods, namely Conditional Automated Channel Pruning (CACP)~\cite{CACP2021IEEESPL}, and GhostNet~\cite{HanGhostNet2020CVPR}. Our pruned VGG16 model performed better than both state-of-the-art methods. Note that when $\alpha=20$, our pruned model contained 2.38M parameters, which are fewer than the pruned models of both CACP and GhostNet, which contained 4.41M and 3.30M parameters. Despite this, our pruned VGG16 still produced an accuracy change of $+0.27$ which is better than both state-of-the-art methods and our baseline unpruned VGG16 model. Moreover, when $\alpha=10$, our pruned model contained only 1.05M parameters, but it had an accuracy change similar to GhostNet, having 3.30M parameters.

Table~\ref{tab:soa} also shows the comparison of proposed method with four state-of-the-art methods on ResNet18 architecture and CIFAR-10 dataset.
The state-of-the-art methods include Prune it Yourself (PIY)~\cite{YanMIPR2020}, Conditional Automated Channel Pruning (CACP)~\cite{CACP2021IEEESPL}, Lagrangian Relaxation (P-SLR)~\cite{GurevinArxiv2020}, and PCNN~\cite{ZhanhongPCNNDAC2020}. It can be seen that the proposed method outperformed all the methods in terms of accuracy and three out of four state-of-the-art methods in terms of compression.
Table~\ref{tab:soa} also shows the comparison of proposed method with state-of-the-art AMC \cite{he2018amc} method on ResNet50. Our method at $\alpha=55$ outperforms it. 

\begin{table}[t]
  \centering
  \caption{Comparison of proposed Constrained Reinforcement Learning (CRL) method with state-of-the-art approaches on VGG16~\cite{simonyan2015vgg}, ResNet18~\cite{he2015resnet} and ResNet50~\cite{he2015resnet} architectures using CIFAR-10 dataset.}
  \resizebox{0.99\columnwidth}{!}{
    \begin{tabular}{c|l|cc|ccc}
    \multicolumn{1}{r}{} &       & \multicolumn{2}{c|}{ \textbf{Unpruned}} & \multicolumn{3}{c}{ \textbf{Pruned}} \\
\cmidrule{3-7}    \multicolumn{1}{r}{} &       & \multicolumn{1}{c}{ \textbf{Acc. $\%$}} & \multicolumn{1}{c|}{ \textbf{Params.}} & \multicolumn{1}{c}{ \textbf{Acc. $\%$}} & \multicolumn{1}{c}{ \textbf{Params.}} & \multicolumn{1}{c}{ \textbf{$\Delta$ Acc. $\%$ }} \\
\multicolumn{1}{r}{} &       & \multicolumn{1}{c}{ } & \multicolumn{1}{c|}{ {in millions}} & \multicolumn{1}{c}{ } & \multicolumn{1}{c}{ in millions} & \multicolumn{1}{c}{} \\
\hline
    \multirow{4}[1]{*}{\rotatebox[origin=c]{90}{VGG16}} & {CACP~\cite{CACP2021IEEESPL}} & 93.02 & 14.73 & 92.89 & \textcolor{red}{4.41}  & -0.13 \\
          & {Ghost Net~\cite{HanGhostNet2020CVPR}} & 93.60 & 14.73 & 92.90 & 3.30  & -0.70 \\
          & {CRL-$\alpha = 20\%$} & 90.69 & 14.73 & 90.96 & {2.38} & \textbf{0.27} \\
          & {CRL-$\alpha = 10\%$} & 90.69 & 14.73 & 89.89 & \textbf{1.05} & \textcolor{red}{-0.80} \\
    \midrule
    \multirow{6}[2]{*}{\rotatebox[origin=c]{90}{ResNet18}} & {PIY~\cite{YanMIPR2020}} & 91.78 & 11.18 & 91.23 & \textcolor{red}{6.11}  & -0.55 \\
          & {CACP~\cite{CACP2021IEEESPL}} & 93.02 & 11.68 & 92.03 & 3.50   & -0.99 \\
          & {P-SLR~\cite{GurevinArxiv2020}} & 93.33 & 11.68 & 90.37 & \bf{1.34}  & \textcolor{red}{-2.96} \\
          & {PCNN~\cite{ZhanhongPCNNDAC2020}} & 96.58 & 11.20  & 96.38 & 3.80   & -0.20 \\
          & {CRL-$\alpha = 30\%$} & 91.82 & 11.68 & 92.09 & 2.91  & \bf{0.27} \\
          & {CRL-$\alpha = 20\%$} & 91.82 & 11.68 & 90.97 & 2.52  & -0.85 \\
    \midrule
    \multirow{2}[2]{*}{\rotatebox[origin=c]{90}{ResNet50}}
          & {AMC~\cite{he2018amc}} & 93.53 & 25.56 & 93.55 & 15.34   & 0.02 \\
          & {CRL-$\alpha = 55\%$} & 92.97 & 25.56 & 93.60 & \bf{12.34}  & \bf{0.63} \\          
              
    
    \end{tabular}
    }
  \label{tab:soa}%
\end{table}%


\section{Conclusions}
\label{sec:concl}
\vspace{-0.2cm}

We propose a novel framework for neural network pruning via constrained reinforcement learning that allows respecting budgets on arbitrary, possibly non-differentiable functions. Ours is a pro-Lagrangian approach that incorporates budget constraints by constructing a trust region containing all policies that respect constraints. Our experiments show that the proposed CRL strategy significantly outperform the state-of-the-art methods in terms of producing small and compact while maintaining the accuracy of unpruned baseline architecture. Specifically, our method reduces nearly $75.08\%-92.9\%$ parameters without incurring any significant loss in performance.



\newpage
\small
\bibliographystyle{IEEEbib}
\bibliography{egbib}

\end{document}